\documentclass[11pt,a4paper]{article}
\usepackage[margin=1in]{geometry}
\usepackage{fancyheadings}
\usepackage{emnlp2017}
\usepackage{times}
\usepackage{latexsym}
\usepackage{multirow}
\usepackage{url}
\usepackage{graphicx}
\usepackage{subcaption}
\usepackage{amsmath}
\usepackage{mdframed}

\usepackage{enumitem}

\usepackage{etoolbox}
\providetoggle{long}
\settoggle{long}{false}

\setitemize{noitemsep,topsep=0pt,parsep=0pt,partopsep=0pt,leftmargin=3pt,rightmargin=1pt}

\usepackage{tabu}
\emnlpfinalcopy

\def\emnlppaperid{***}

\title{Controlling Linguistic Style Aspects in Neural Language Generation}
\author{Jessica Ficler \and Yoav Goldberg \\
 Computer Science Department \\
 Bar-Ilan University \\
 Israel \\
 {\tt \{jessica.ficler, yoav.goldberg\}@gmail.com} 
}
\date{}

\begin{document}
\maketitle
\rhead{}
\chead{}

\lhead{Accepted as a long paper in  Stylistic Variation Workshop 2017}

\thispagestyle{fancy}

\begin{abstract}
Most work on neural natural language generation (NNLG) focus on controlling the content of the generated text. We experiment with controlling several stylistic aspects of the generated text, in addition to its content. The method is based on conditioned RNN language model, where the desired content as well as the stylistic parameters serve as conditioning contexts.
We demonstrate the approach on the movie reviews domain and show that it is successful in generating coherent sentences corresponding to the required linguistic style and content.

\end{abstract}

\section{Introduction}
The same message (e.g. expressing a positive sentiment towards the plot of a movie) can be conveyed in different ways. It can be long or short, written in a professional or colloquial style, written in a personal or impersonal voice, and can make use of many adjectives or only few.

Consider for example the following to sentences:

\vspace{5pt}
\noindent
(1) \textit{``A genuinely unique, full-on sensory experience that treads its own path between narrative clarity and pure visual expression."}\\ 
(2) \textit{``OMG... This movie actually made me cry a little bit because I laughed so hard at some parts lol."}
\vspace{5pt}

They are both of medium length, but the first appears to be written by a professional critic, and uses impersonal voice and many adjectives; while the second is written in a colloquial style, using a personal voice and few adjectives.

In a text generation system, it is desirable to have control over such stylistic aspects of the text: style variations are used to express the social meanings of a message, and controlling the style of a text is necessary for appropriately conveying a message in a way that is adequate to the social context 
\cite{biber2009register,niederhoffer2002linguistic}. This work focuses on generating text while allowing control of its stylistic properties.

The recent introduction of recurrent neural language models and recurrent sequence-to-sequence
architectures to NLP brought with it a surge of work on natural language
generation. Most of these research efforts focus on controlling the \emph{content} of
the generated text \cite{lipton2015capturing,choiglobally,lebretneural,choiglobally,tang2016context,radford2017learning}, while 
a few model more stylistic aspects of the generated text such as the identity of the speaker in a
dialog setting \cite{li2016persona}; the politeness of the generated
text or the text length in a machine-translation setting
\cite{sennrich2016controlling,kikuchi2016controlling}; or the tense in generated movie reviews \cite{hu2017controllable}. Each of these works targets a single, focused stylistic aspect of the text. \emph{Can we achieve finer-grained control over the generated outcome, controlling several stylistic aspects simultaneously?}

We explore a simple neural natural-language generation (NNLG) framework
that allows for high-level control on the generated content (similar to previous work) as well as control over multiple stylistic
properties of the generated text.
We show that we can indeed achieve control over each of the individual properties.

As most recent efforts, our model (section \ref{sec:proposal}) is based on a conditioned language model, in which the generated
text is conditioned on a context vector.\footnote{ See \cite{hoang2016incorporating} for other conditioning models.}
In our case, the context vector encodes
a set of desired properties that we want to be present in the generated text.\footnote{Another view is that of an
encoder-decoder model, in which the encoder component encodes the set of desired
properties.} At training time, we work in a fully supervised setup, in which each sentence is labeled with a set of linguistic properties we want to condition on. These are encoded into the context vector, and the model is trained to generate the sentence based on them. At test time, we can set the values of the individual properties to get the desired response.
As we show in section
\ref{sec:eval-generalize}, the model generalizes fairly well, allowing the
generation of text with property combinations that were not seen during training.

The main challenge we face is thus obtaining the needed annotations for training time. In section \ref{sec:data-set} we show how such annotations can be obtained from meta-data or using specialized text-based heuristics.

Recent work \cite{hu2017controllable} tackles a similar goal to ours. They propose a novel generative model combining variational auto-encoders and holistic attribute discriminators, in order to achieve individual control on different aspects of the generated text. Their experiments condition on two aspects of the text (sentiment and tense), and train and evaluate on sentences of up to 16 words. In contrast, we propose a much simpler model and focus on its application in a realistic setting: we use all naturally occurring sentence lengths, and generate text according to two content-based parameters (sentiment score and
topic) and four stylistic parameters (the length of the text, whether it is 
descriptive, whether it is written in a personal voice, and whether it is written in professional style).
Our model is based on a well-established technology - conditioned language models that are based on Long Short-Term Memory (LSTM), which was proven as strong and effective sequence model.

We perform an extensive evaluation, and verify that the model indeed learns to associate the different
parameters with the correct aspects of the text, and is in many cases able to generate
sentences that correspond to the requested parameter values.
We also show that conditioning on the given properties in a conditioned language model
indeed achieve better perplexity scores compared to an unconditioned language model
trained on the entire dataset, and also compared to unconditioned models that
are trained on subsets of the data that correspond to a particular conditioning
set.
Finally, we show that the model is able to generalize, i.e., to generate sentences for
combinations that were not observed in training. 

\begin{table*}[t!]
\scalebox{0.77}{
\centering
{\tabulinesep=1.5mm
\begin{tabu}{p{0.3cm}|p{1.8cm}|p{5cm}|p{1.5cm}|c|p{7.5cm}} 
&\bf Parameter &\bf Description &\bf Source &\bf Possible values & \bf Examples \\
\hline
\multirow{10}{*}{\rotatebox[origin=c]{90}{Style}}&\multirow{2}{*}{\parbox[t]{1,5cm}{Professional}} & \multirow{2}{*}{\parbox[t]{5cm}{Whether the review is written in the style of a professional critic or not}} & \multirow{2}{*}{\parbox[t]{2cm}{meta-data}}
& False &
\parbox[t]{7.5cm}
{``So glad to see this movie !!"}
\\
\cline{5-6}
&&& &True &
\parbox[t]{7.5cm}
{``This is a breath of fresh air, it's a welcome return to the franchise's brand of satirical humor."}
\\
\cline{2-6}

&\multirow{2}{*}{\parbox[t]{1.9cm}{Personal}} &
\multirow{2}{*}{\parbox[t]{5cm}{Whether the review describes subjective
experience (written in personal voice) or not}} &\multirow{2}{*}{\parbox[t]{2cm}{content derived}} &  False &\parbox[t]{7.5cm}
{``Very similar to the book."}
\\
\cline{5-6}
&&& &True &\parbox[t]{7.5cm}
{``I could see the movie again, ``The Kid With Me'' is a very good film."}
\\
\cline{2-6}

&\multirow{2}{*}{Length} & \multirow{2}{*}{\parbox[t]{5cm}{Number of words}}
&\multirow{2}{*}{\parbox[t]{2cm}{content derived}} & \multicolumn{2}{l}{ $\leq$
10 words / 11-20 words / 21-40 words / $>$ 40 words } \\
&&& \\
\cline{2-6}

&\multirow{2}{*}{Descriptive} &\multirow{2}{*}{\parbox[t]{5cm}
{Whether the review is in descriptive style or not}
} &\multirow{2}{*}{\parbox[t]{2cm}{content
derived}} &  True &\parbox[t]{7.5cm}
{``Such a hilarious and funny romantic comedy."}\\
\cline{5-6}
&&& &  False &\parbox[t]{7.5cm}
{``A definite must see for fans of anime fans, pop culture references and animation with a good laugh too."}
\\
\hline

\multirow{10}{*}{\vspace{3cm}\rotatebox{90}{Content}}&\multirow{2}{*}{\parbox[t]{2cm}{Sentiment}}
& \multirow{2}{*}{\parbox[t]{5cm}{The score that the reviewer gave the
movie}} &\multirow{2}{*}{\parbox[t]{2cm}{meta-data}} & Positive & \parbox[t]{7.5cm}
{``In other words: ``The Four'' is so much to keep you on the edge of your seat."}
\\
\cline{5-6}
&&& &  Neutral &\parbox[t]{7.5cm}
{``While the film doesn't quite reach the level of sugar fluctuations, it's beautifully animated."}
\\
\cline{5-6}
&&& & Negative &\parbox[t]{7.5cm}
{``At its core ,it's a very low-budget movie that just seems to be a bunch of fluff."}
\\
\cline{2-6}

&\multirow{2}{*}{Theme} &
\multirow{2}{*}{\vspace{-0.5cm}\parbox[t]{4.8cm}{Whether the sentence's content is
about the \emph{Plot}, \emph{Acting}, \emph{Production}, \emph{Effects} or none
of these (\emph{Other})}} &\multirow{2}{*}{\parbox[t]{2cm}{content derived}}&  Plot&\parbox[t]{7.5cm}
{``The characters were great and the storyline had me laughing out loud at the beginning of the movie."}
\\
\cline{5-6}
&&& &  Acting&\parbox[t]{7.5cm}
{``The only saving grace is that the rest of the cast are all excellent and the pacing is absolutely flawless."}
\\
\cline{5-6}
&&& &  Production&\parbox[t]{7.5cm}
{``If you're a Yorkshire fan, you won't be disappointed, and the director's magical."}
\\
\cline{5-6}
&&& &  Effects&\parbox[t]{7.5cm}
{``Only saving grace is the sound effects."}
\\
\cline{5-6}
&&& &  Other&\parbox[t]{7.5cm}
{``I'm afraid that the movie is aimed at kids and adults weren't sure what to say about it."}
\\
\hline

\end{tabu}
}}
\caption{Parameters and possible values in the movie-reviews domain.}
\label{tbl:params}
\end{table*}

\section{Task Description and Definition}

Our goal is to generate natural language text that conforms to a set of content-based and stylistic properties. The generated text should convey the information requested by the content properties, while conforming to the style requirements posed by the style properties.

For example in the movie reviews domain, \texttt{theme} is a content parameter
indicating the topical aspect which the review refers to (i.e. the plot, the
acting, and so on); and  \texttt{descriptive} is a style parameter that
indicates whether the review text uses many adjectives.
The sentence 
\textit{``A wholly original, well-acted, romantic comedy that's elevated by the modest talents of a lesser known cast."}
corresponds to \texttt{theme:acting} and
\texttt{descriptive:true}, as it
includes many descriptions and refers to the acting, while the sentence
\textit{``In the end, there are some holes in the story, but it's an exciting and tender film."} corresponds to \texttt{theme:plot} and \texttt{descriptive:false}.

More formally, we assume a set of $k$ parameters $\{p_1,\dots, p_{k}\}$, each
parameter $p_i$ with a set of possible values $\{v_1,\dots, v_{p_i}\}$. Then,
given a specific assignment to these values our goal is to generate a text that is compatible with the parameters values. 
Table \ref{tbl:params} lists the full set of parameters and values we
consider in this work, all in the movie reviews domain. In section \ref{sec:data-set} we discuss in detail the different parameters and how we
obtain their values for the texts in our reviews corpus.

To give a taste of the complete task, we provide two examples of possible value assignments and sentences corresponding to them:
\begin{center}
\scalebox{0.8}{
\begin{tabular}{l|l||l|l} 
Type&Parameter & Value (1) &Value (2) \\
\hline
Content&Theme & Acting   & Other \\
Content&Sentiment &  Positive & Negative\\ 
Style&Professional &  True   & False \\
Style&Personal & False  & True \\
Style&Length & 21-40 words   &11-20 words \\
Style&Descriptive & False  & True \\
\end{tabular}}
\end{center}

\noindent \underline{Sentences for value set 1:}
\begin{itemize}
\item ``This movie is excellent, the actors aren't all over the place ,but the movie has a lot of fun, exploring the lesson in a way that they can hold their own lives."
\item ``It's a realistic and deeply committed performance from the opening shot, the movie gives an excellent showcase for the final act, and the visuals are bold and daring."
\iftoggle{long}{\item ``It's not an easy task but the cast is top-notch, the visuals aren't particularly as good as any other film, but it 's still a must see for any actor."}{}
\end{itemize}

\vspace{0.2cm}

\noindent \underline{Sentences for value set 2:}
\begin{itemize}
\item ``My biggest gripe is that the whole movie is pretty absurd and I thought it was a little too predictable."
\item ``The first half is pretty good and I was hoping for a few funny moments but not funny at all."
\iftoggle{long}{\item ``My biggest problem with the whole movie though is that there is nothing new or original or great in this film."}{}
\end{itemize}

\section{Conditioned Language Model}
\label{sec:proposal}

Like in previous neural language-generation work
\cite{lipton2015capturing,tang2016context},
our model is also a conditioned language model. In a regular language model (LM), each
token $w_t$ is conditioned on the previous tokens, and the probability of a
sentence $w_1,...,w_n$ is given by:
\begin{equation}
P(w_1,...,w_n) = \Pi_{t=1}^n P(w_t | w_1, \dots w_{t-1})
\end{equation}
In a conditioned language model, we add an additional conditioning context, $c$:
\begin{equation}
P(w_1,...,w_n|c) = \Pi_{t=1}^n P(w_t | w_1, \dots w_{t-1}, c)
\end{equation}
Each token in the sentence is conditioned on the previous ones, as well the additional context $c$.

A conditioned language model can be implemented using an recurrent neural
network language model (RNN-LM, \cite{mikolov2010recurrent}), where the context
$\mathbf{c}$ is a vector that is concatenated to the input vector at each time step.

Conditioned language models were shown to be effective for natural language
generation. We differ from previous work by the choice of
conditioning contexts, and by conditioning on many parameters simultaneously.

In our case, the condition vector $\mathbf{c}$ encodes the desired textual
properties. Each parameter value is associated with an embedding vector, and $\mathbf{c}$ is a concatenation of these embedding vectors. The vector $\mathbf{c}$ is fed into the RNN at each step, concatenated to
the previous word in the sequence.

\paragraph{Technical Details} We use an LSTM-based language model
\cite{hochreiter1997long}, and encode the vocabulary using Byte Pair Encoding (BPE), which allows representation of an open vocabulary through
a fixed-size vocabulary by splitting rare words into subword units, providing a convenient way of dealing with rare words.
Further
details regarding layer sizes, training regime, vocabulary size and so on are
provided in the supplementary material.

\section{Data-set Collection and Annotation}
\label{sec:data-set}
For training the model, we need a dataset of review texts, each annotated with 
a value assignment to each of the style and the content parameters.
We obtain these values from two sources:
(1) We derive it from meta-data associated with the review, when available. (2)
We extract it from the review text using a heuristic. We use three kinds
of heuristics: based on lists of content-words; based on the existence of
certain function words; and based on the distribution on part-of-speech tags.
These annotations may contain noise, and indeed some of our heuristics are not very tight. We demonstrate that we can achieve good performance despite the noise. Naturally, improving the heuristics is likely to results in improved performance.

Our reviews corpus is based on the Rotten-Tomatoes
website.\footnote{\url{http://www.rottentomatoes.com}}
We collected 1,002,625 movie reviews for 7,500 movies and split them into sentences. Each sentence is then annotated according to four style parameters (\texttt{professional}, \texttt{personal},
\texttt{descriptive} and \texttt{length}) and two content parameters
(\texttt{sentiment} and \texttt{theme}). The meanings of these properties and how we obtain values for them are described below.

\subsection{Annotations Based on Meta-data}
\paragraph{Professional} indicates whether the review is written in a  professional (\texttt{true}) or a colloquial (\texttt{false}) style. We label sentences as \texttt{professional:true} if it is written by either (1) a reviewer that is a professional critic; (2) a reviewer that is marked as a ``super-reviewer'' on the RottenTomatoes website (a title given to reviewers who write high-quality reviews). Other sentences are labeled as \texttt{professional:false}.
\paragraph{Sentiment} reflects the grade that was given by the review writer.
Possible values
for grade are: \texttt{positive}, \texttt{neutral}, \texttt{negative} or
\texttt{none}. In audience reviews the movies are rated by the reviewer on a
scale of 0 to 5 stars. In critic reviews, the score was taken from the original
review (which is external to the rotten-tomatoes website). We normalized the
critics scores to be on 0-5 scale. We then consider reviews with grade 0-2 as \texttt{negative}, 3 as \texttt{neutral} and 4-5 as \texttt{positive}.
Cases where no score information was available are labeled as \texttt{none}.\footnote{Note that while the sentiment scores are assigned to a complete review, we associate them here with individual sentences. This is a deficiency in the heuristic, which may explain some of the failures observed in section \ref{sec:individual}.}

\begin{figure*}
\centering
\scalebox{0.9}{
{\tabulinesep=0.2mm
\begin{tabu}{cccccc} 
\small \rotatebox{90}{(a) Professional} &
\includegraphics[scale=0.5]{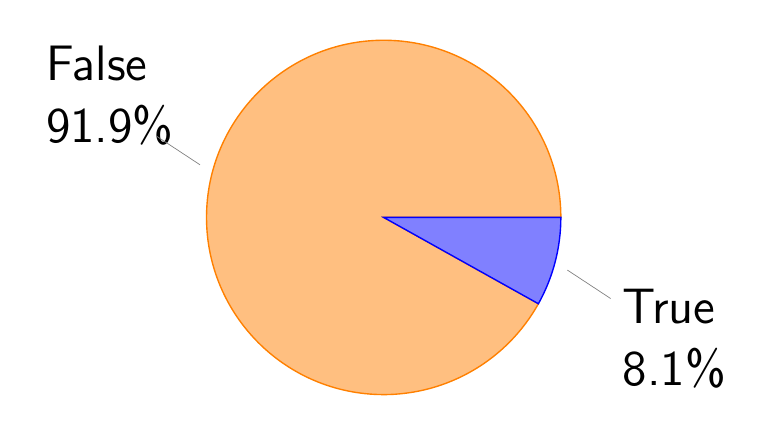}&
\small \rotatebox{90}{(b) Personal} &
\includegraphics[scale=0.5]{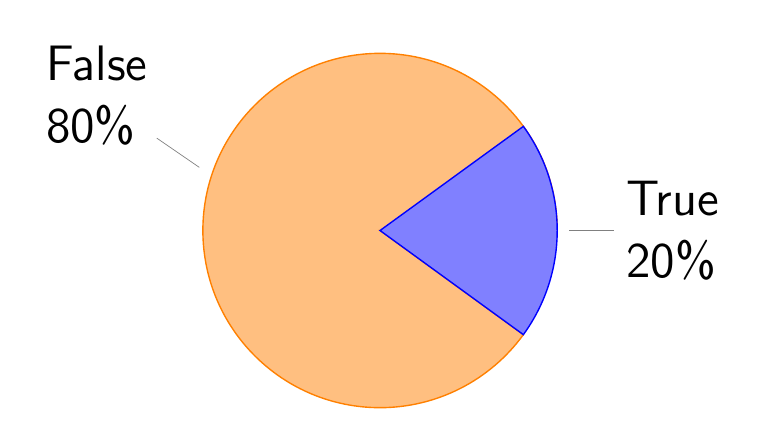}&
\small \rotatebox{90}{(c) Descriptive} &
 \includegraphics[scale=0.5]{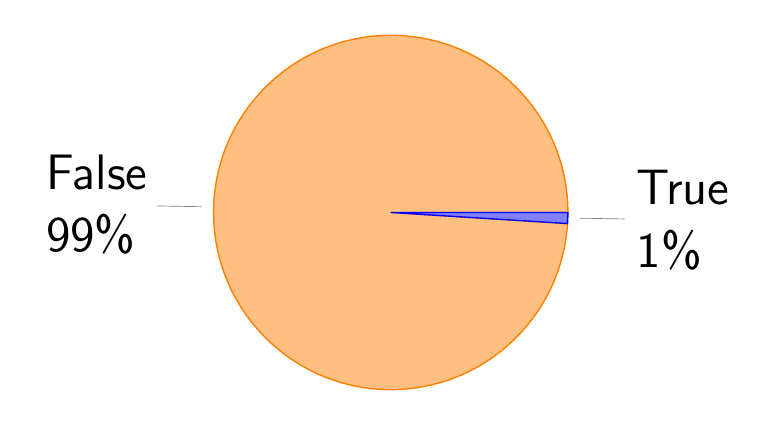}\\
 \small \rotatebox{90}{(d) Sentiment} &
\includegraphics[scale=0.5]{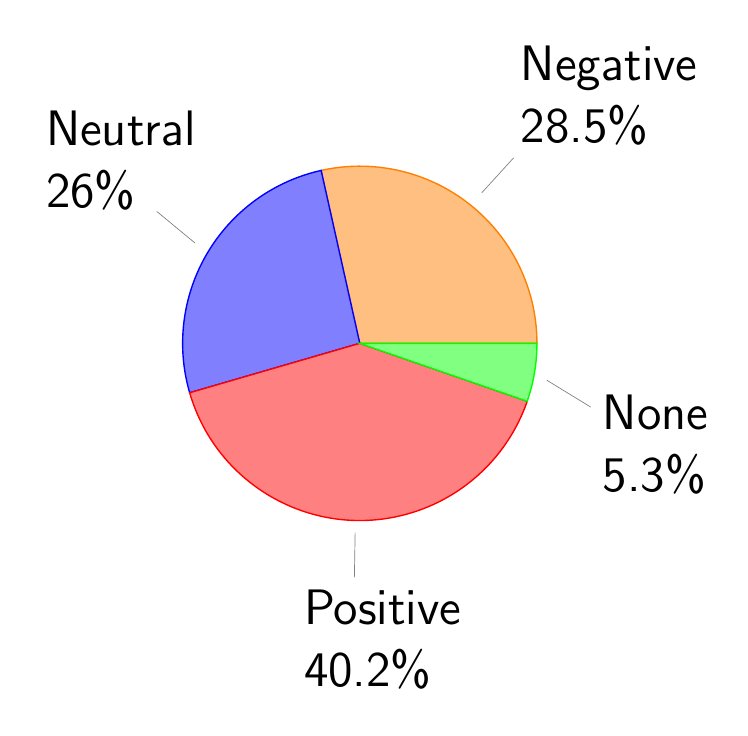}&
 \small \rotatebox{90}{(e) Theme} & \hspace{0.4cm}
\includegraphics[scale=0.5]{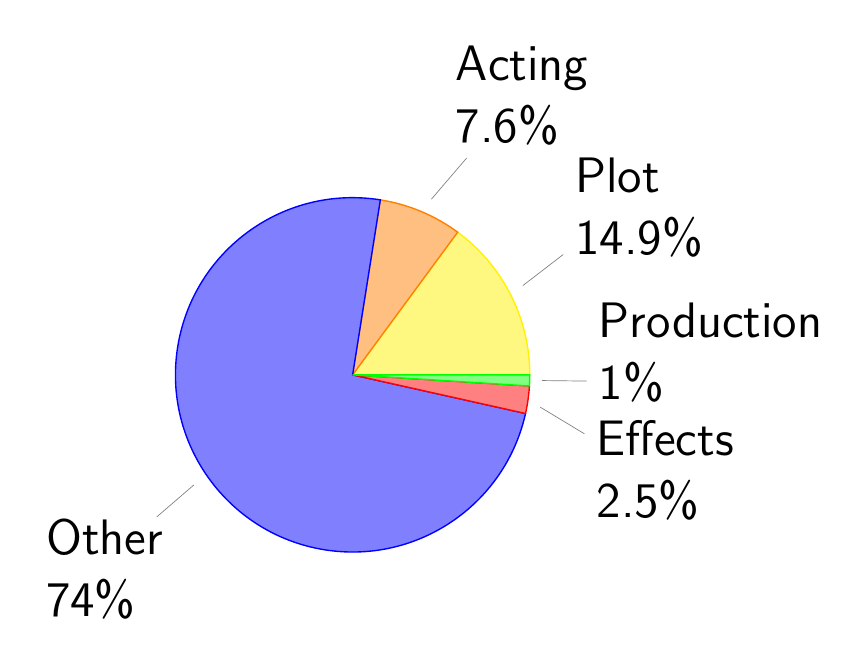}&
 \small \rotatebox{90}{(f) Length} &
\includegraphics[scale=0.48]{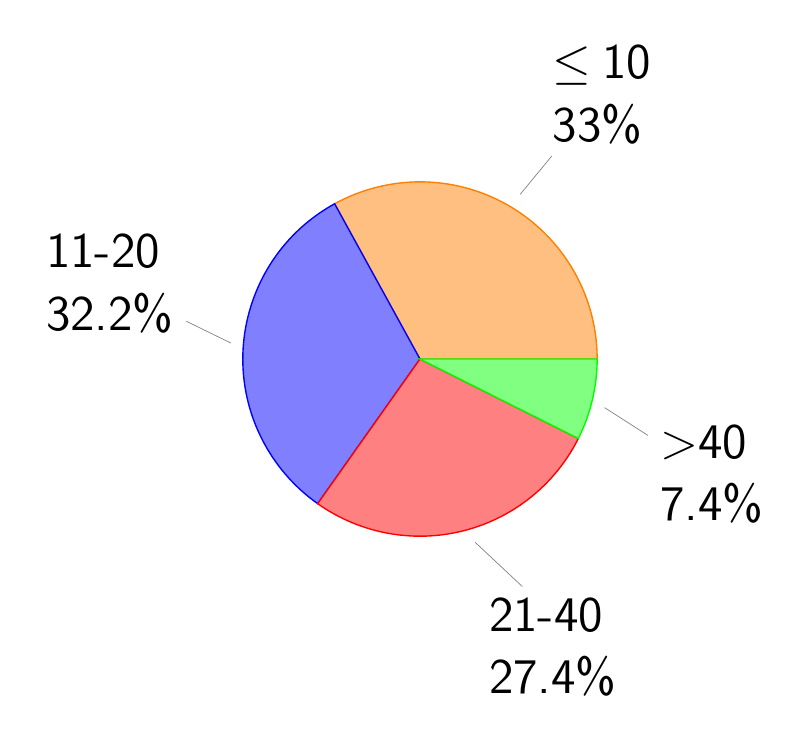} 
\end{tabu}}}
\caption{Movie reviews data-set statistics.}
\label{fig:data-stat}
\end{figure*}

\subsection{Annotations Derived from Text}
\paragraph{Length} We count the number of tokens in the sentence and
associate each sentence to one of four bins: $\leq$10, 11-20, 21-40, $>$40.

\paragraph{Personal} whether the sentence is written in a personal voice, indicating a subjective point of view
(\textit{``I thought it was a good movie."}, \textit{``Just not my cup of tea."}) or not (\textit{``Overall, it is definitely worth watching."}, \textit{``The movie doesn't bring anything new."}),
We label sentences that include the personal pronoun or
possessive (\textit{``I''}, \textit{``my''}) as \texttt{personal:true} and
others as \texttt{personal:false}.
\paragraph{Theme} the aspect of the movie that the sentence refers to. The possible values are \texttt{plot}, \texttt{acting}, \texttt{production} and \texttt{effects}. 
We assign a category to a sentence using word lists. We went over the 
frequent words in the corpus, and looked for words that we believe are
indicative of the different aspects (i.e., for \texttt{plot} this includes words
such as \emph{sciprt, story, subplots}. The complete word lists are available in
the supplementary material).
Each sentence was labeled with the category that has the most words in the
sentence. Sentences that do not include any words from our lists are labeled as \texttt{other}.

\paragraph{Descriptive} whether the sentence is descriptive
(\textit{``A warm and sweet, funny movie."}\iftoggle{long}{
, 
\textit{``A little bit of a predictable and boring romantic comedy with a few funny moments but overall pretty entertaining."}}{}) or not (\textit{``It's one of the worst movies of the year, but it's not a total waste of time."}\iftoggle{long}{
, \textit{``If you are a fan of the first movie, this is a lot of fun to watch."}}{}),
Our (somewhat simplistic)
heuristic is based on the
premise that descriptive texts make heavy use of adjectives. We labeled a
sentence as \texttt{descriptive:true} if at least 35\% of its part-of-speech
sequence tags are adjectives (JJ).
All other sentences were considered as non-descriptive.

\subsection{Dataset Statistics}
Our final data-set includes 2,773,435 sentences where each sentence is labeled with the 6 parameters.
We randomly divided the data-set to training (\#2,769,138), development  (\#2,139) and test (\#2,158) sets.
Figure \ref{fig:data-stat} shows the distribution of the different properties
in the dataset.

\section{Evaluating Language Model Quality}

In our first set of experiments, we measure the quality of the conditioned
language model in terms of test-set perplexity. 

\subsection{Conditioned vs. Unconditioned}
Our model is a language model that is conditioned on various parameters.
As a sanity check, we verify that knowing the parameters indeed helps in
achieving better language modeling results. We compare the dev-set and test-set 
perplexities of our conditioned language model to an unconditioned (regular)
language model trained on the same data.  
The results, summarized in the following table, show that knowing the correct parameter values indeed results in better perplexity.
\begin{table}[h!]
\begin{center}
\scalebox{1}{
\begin{tabular}{l|c|c} 
& dev& test\\
\hline
Not-conditioned & 25.8&  24.4\\
Conditioned & \bf 24.8& \bf 23.3\\
\end{tabular}}
\caption{Conditioned and not-conditioned language model perplexities on the development and test sets.}
\end{center}
\end{table}

\subsection{Conditioned vs. Dedicated LMs}
A second natural baseline to the conditioned LM is to train a
separate unconditioned LM on a subset of the data. For example, if we are
interested in generating sentences with the properties
\texttt{personal:false}, \texttt{sentiment:pos},
\texttt{professional:false}, \texttt{theme:other} and \texttt{length:$\leq$10}, we will train a dedicated LM on just the sentences that fit these
characteristics. 

We hypothesize that the conditioned LM trained on all the data will
be more effective than a dedicated LM, as it will be able to generalize across
properties-combinations, and share data between the different
settings.
In this set of experiment, we verify this hypothesis. 

For a set of parameters and values $\{p_1, p_2, \cdots p_n\}$, we train $n$
sub-models where each sub-model $m_i$ is trained on the subset of sentences that
match parameters $\{p_1, p_2, \cdots p_i\}$.
For example, given the set of parameters values as above,  
we train 5 sub-models: the first on data with \texttt{personal:false} only,
the second on data with \texttt{persoal:false} and
\texttt{sentiment:positive}, etc.  As we add parameters, the size of the
training set of the sub-model decreases.

\begin{figure*}[t!]
\centering
\scalebox{0.84}{
\begin{tabu}{cccc} 
 \includegraphics[width=0.26\textwidth]{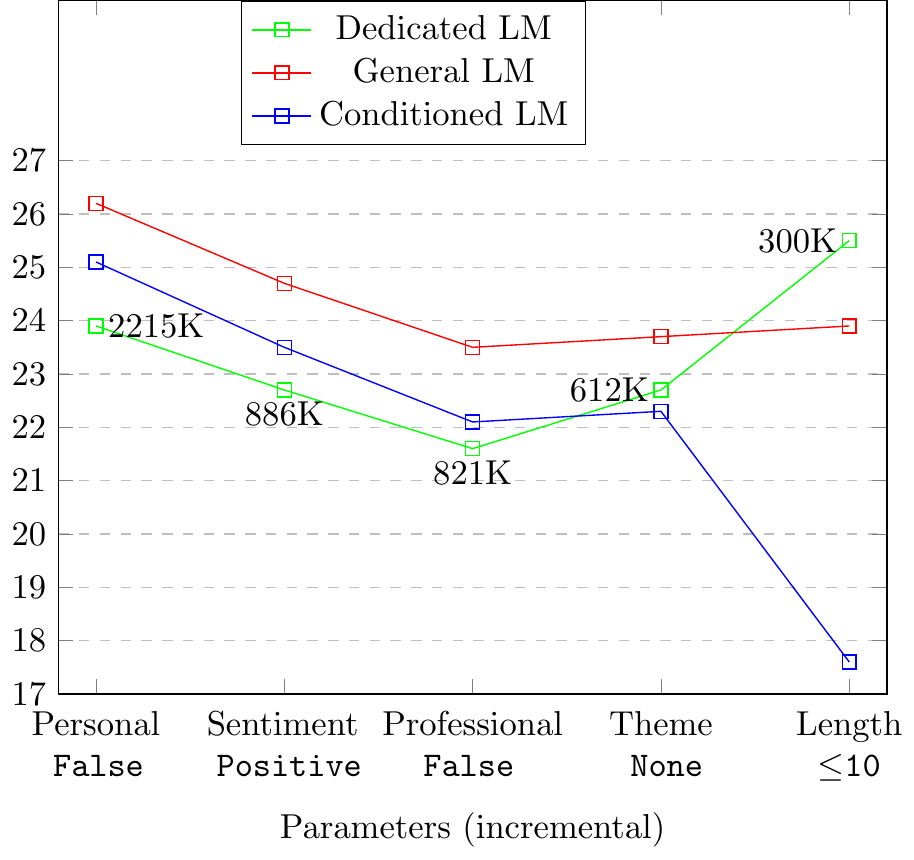} &
\includegraphics[width=0.26\textwidth]{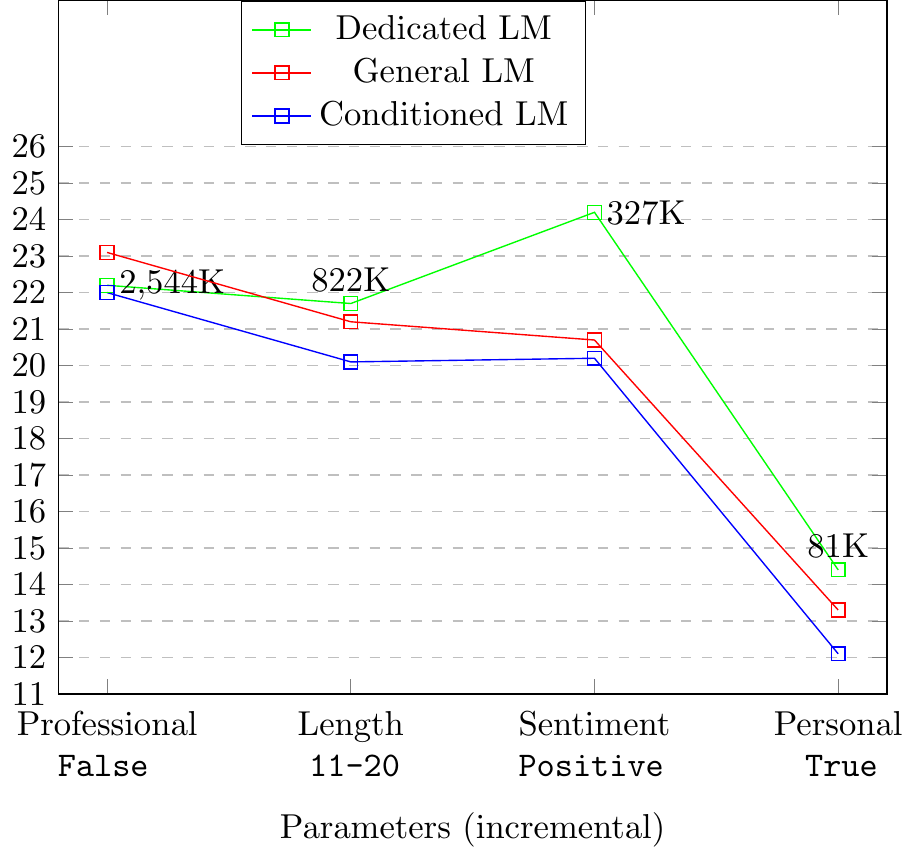}  &
\includegraphics[width=0.26\textwidth]{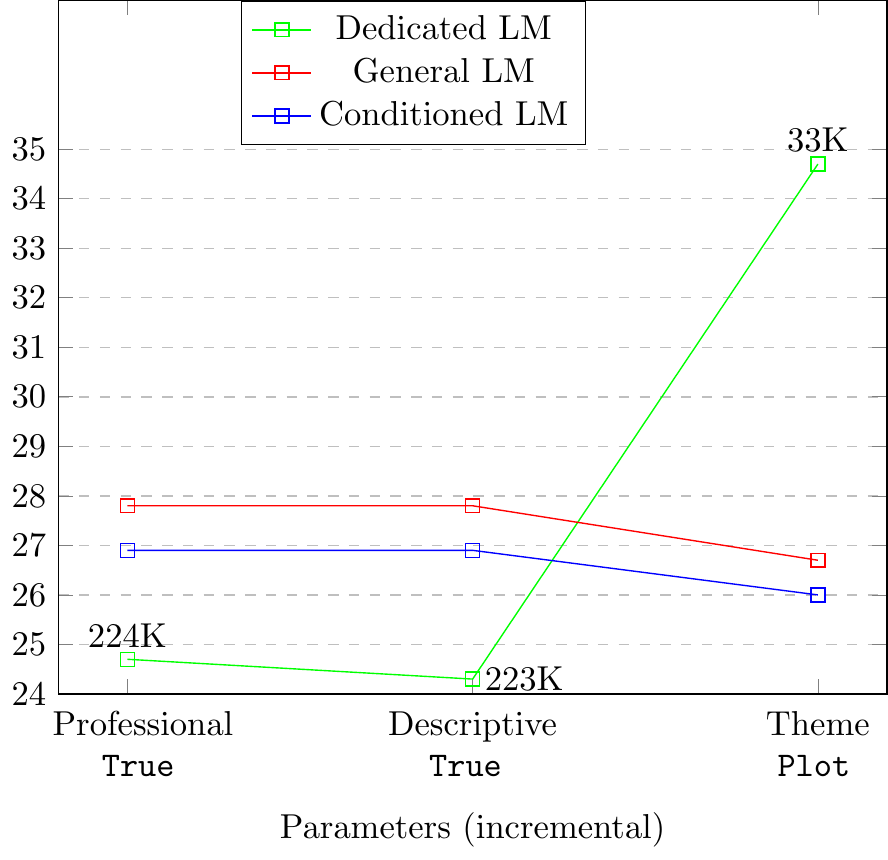}  & 
\includegraphics[width=0.26\textwidth]{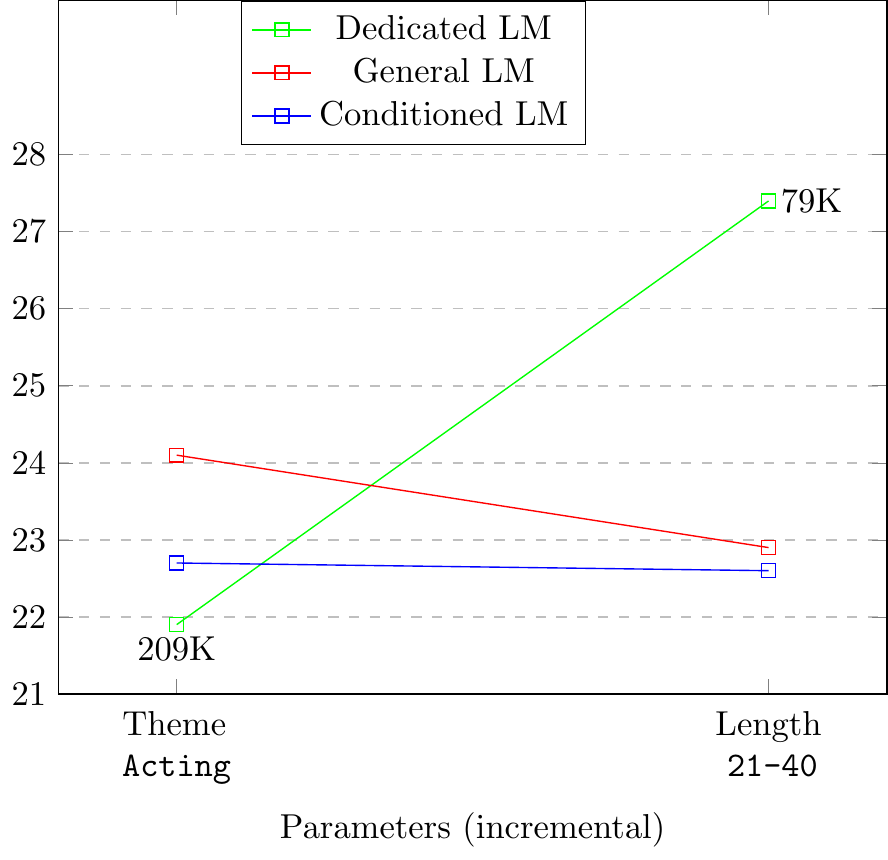}  \\
(a) & (b) & (c) & (d) \\
\end{tabu}}
\caption{Perplexities of conditioned, unconditioned and dedicated language
models for various parameter combinations. The numbers on the dedicated-model line indicates the number of sentences that the sub-model was trained on.}
\label{fig:parts}
\end{figure*}

For each dedicated sub-model, we measure its perplexity on the test-set sentences that
match the criteria, and compare it to a conditioned LM with these
criteria, and to an unconditioned language model.
We do this for 4 different parameter-sets. Figure \ref{fig:parts} presents the
results.

The results indicate that when only few conditioning parameters are needed,
and if the coverage of the parameter combination in the training set is large
enough, the dedicated LM approach indeed outperforms the conditioned LM. This is
the case in the first three sub-models in \ref{fig:parts}a, and the first two
sub-models in \ref{fig:parts}c. With few conditioning criteria, the dedicated LM approach is effective. However, it is not scalable. As we increase the number of
conditioning factors, the amount of available training data to the dedicated
model drops, and so does the modeling quality.
In contrast, the conditioned model manages to generalize from sentences with
different sets of properties, and is effective also with large number of
conditioning factors. We thus conclude that for our use case, in which we need
to condition on many different aspects of the generated sentence, the
conditioned LM is far more suitable than the dedicated LM.

\subsection{Conditioned vs. Flipped Conditioning}
\label{sec:flipping}
The previous experiments show that a conditioned model outperforms an
unconditioned one. Here, we focus on the effect of the individual conditioning
parameters.
We compare the perplexity when using the correct conditioning values to the
perplexity achieved when flipping the parameter value to an incorrect one. We do
that for parameters that have opposing values: \texttt{personal},
\texttt{professional}, \texttt{sentiment} and \texttt{descriptive}. 
The following table summarizes the results:

\begin{table}[h!]
\begin{center}
\scalebox{0.9}{
\centering
\begin{tabular}{l|c} 
Correct Value&23.3\\
\hline
Replacing Descriptive with non-Descriptive & 27.2\\
Replacing Personal & 27.5\\
Replacing Professional & 25\\
Replacing Sentiment \texttt{Pos} with \texttt{Neg} & 24.3\\
\end{tabular}}
\end{center}
\caption{Test-set perplexities when supplying the correct parameter values and
when supplying the opposite values.}
\end{table}
\noindent
There is a substantial drop in quality
(increase in perplexity) when flipping the parameter values.
The drop is smallest for sentiment, and largest for descriptiveness and
personal voice. We conclude that the model distinguishes descriptive text and personal voice better than it distinguishes sentiment and professional text.

\section{Evaluating the Generated Sentences}
In section \ref{sec:flipping} we verified the effectiveness of the conditioned
model by showing that flipping a conditioning parameter value results
in worse perplexity. However, we still need to verify that the model indeed
associates each parameter with the correct behavior. In this set of experiments,
we use the model to generate random sentences with different conditioning
properties, and measure how well they match the requested behavior.

We generated 3,285 sentences according to the following protocol: for each
property-combination attested in the development set, we generated 1,000 random
sentences conditioned on these properties. We then sorted the generated
sentences according to their probability, and chose the top $k=(c_f/m_f)*100$ sentences, 
where $c_f$ is the frequency of the property-combination in the dev set and $m_f$ is the frequency of the most frequent property-combination in the dev set.

This process resulted in 3,285 high-scoring but diverse sentences, with properties
that are distributed according to the properties distribution in the development
set.

\subsection{Capturing Individual Properties}
\label{sec:individual}

\paragraph{Length}
We measure the average, minimum and maximum lengths, and deviation of
the sentences that were generated for a requested length value. 
The following table summarizes the results:
\begin{table}[h!]
\begin{center}
\scalebox{0.85}{
\begin{tabular}{l|c|c|c|c} 
Requested Length& Avg& Min & Max & Deviation$_{m=2}$ \\
\hline
\texttt{<=10} & 7.6 & 1 & 21 & 0.2 \%\\
\texttt{11-20} &20.6 & 5 & 25 & 2.6 \%\\
\texttt{21-40} & 34 & 7 & 49 & 0.6 \%\\
\end{tabular}}
\end{center}
\caption{Average, minimum and maximum lengths of the sentences generated according to the correspond length value; as well as deviation percentage with margin ($m$) of 2.}
\end{table}

The average length fits the required range for each of the values and the percentage of sentences that exceed the limits with margin 2 is between 0.2\% to 2.6\%.

\paragraph{Descriptive}
We measure the percentage of sentences that are considered as descriptive
(containing $>$35\% adjectives) when requesting \texttt{descriptive:true}, and
when requesting \texttt{descriptive:false}.  When requesting descriptive text,
\textbf{85.7\%} of the generated sentences fit the descriptiveness criteria. When requesting
non-descriptive text, \textbf{96\%} of the generated sentences are 
non-descriptive according to our criteria.

\paragraph{Personal Voice}
We measure the percentage of sentences that are considered as personal voice
(containing the pronouns \emph{I} or \emph{my})
when requesting \texttt{personal:true}, and
when requesting \texttt{personal:false}. \textbf{100\%} of the sentence for which
we requested personal voice were indeed in personal voice. When requesting
non-personal text, \textbf{99.85\%} of the sentences are indeed non-personal.

\paragraph{Theme}
For each of the possible theme values, we compute the proportion of the
sentences that were generated with the corresponding value. The confusion matrix
in the following table

shows that the vast majority of sentences are generated according to the
requested theme.\\[0.45em]

\begin{table}[h!]
{\centering
\scalebox{0.64}{
\begin{tabular}{l|c|c|c|c|c} 
Requested value& \% Plot &\% Acting &\% Prod & \% Effects &\% Other \\
\hline
Plot           & 98.7  & 0.8    &0     &0.2    &0.3 \\
Acting         & 2.5  &95.3   &0     &0.6    &1.6\\
Production     & 0     &0       &97.4 &2.6    &0\\
Effects       & 0     &5.9    &0     &91.7   &2.4\\
Other          & 0.04  &0.03    &0     &0.03    &99.9\\
\end{tabular}}
}
\\[0.45em]
\caption{Percentage of generated sentences from each theme, when requesting a given theme value.}
\label{tbl:theme}
\end{table}

\paragraph{Professional}
The \texttt{professional} property of the generated sentences could not be evaluated
automatically, and we thus performed manual evaluation using Mechanical Turk.
We randomly created 1000 sentence-pairs where one is generated with
\texttt{professional:true} and the other with \texttt{professional:false}
(the rest of the property values were chosen randomly). For example in the
following sentence-pair the first is generated with \texttt{professional:true} and the second with \texttt{professional:false}:\\[0.5em]
(t) \textit{``This film has a certain sense of imagination and a sobering look at the clandestine indictment."}\\
(f) \textit{``I know it's a little bit too long, but it's a great movie to watch !!!!"}\\[0.5em]
The annotators were asked to determine which of the sentences was written by a
professional critic. Each of the pairs was annotated by 5 different annotators. When taking a majority vote among the annotators, they were able to tell apart the professional from non-professional sentences generated sentences in \textbf{72.1\%} of the cases. 

When examining the cases where the annotators
failed to recognise the desired writing style,
we saw that in a few
cases the sentence that was generated for \texttt{professional:true} was indeed not
professional enough (e.g. \textit{``Looking forward to the trailer."}, and that in many cases, both sentences could indeed be
considered as either professional or not, as in the following examples:\\[0.5em]
\noindent
(t) \textit{``This is a cute movie with some funny moments, and some of the
jokes are funny and entertaining."}\\
(f) \textit{``Absolutely amazing story of bravery and dedication."}\\[0.5em] 
\noindent
(t) \textit{``A good film for those who have no idea what's going on, but it's a fun adventure."}\\ 
(f) \textit{``An insult to the audience's intelligence."}
\iftoggle{long}{\noindent
(C) \textit{``Just stay away from the end."}\\ 
(A) \textit{``Sloppy script and acting."}\\[0.5em]}{}
\paragraph{Sentiment}
To measure \texttt{sentiment} generation quality, we again perform manual annotations using Mechanical Turk. 
We randomly created 300 pairs of generated sentences for each of the following
settings: \texttt{positive}/\texttt{negative},
\texttt{positive}/\texttt{neutral} and \texttt{negative}/\texttt{neutral}.
The annotators were asked to mark which of the reviewers liked the movie more
than the other. Each of the pairs was annotated by 5 different annotators and we
choose by a majority vote. The annotators correctly identified
\textbf{86.3\%} of the sentence in the Positive/Negative case, \textbf{63\%} of
the sentences in the Positive/Neutral case, and \textbf{69.7\%} of the sentences
in the \texttt{negative}/\texttt{neutral} case. 

Below are some examples for cases where the annotators failed to recognize the intended sentiment:\\[0.5em]
\noindent
(Pos) \textit{``It's a shame that this film is not as good as the previous film, but it still delivers."}\\ 
(Neg) \textit{``The premise is great, the acting is not bad, but the special
effects are so bad."}\\[0.5em]
\noindent
(Pos) \textit{``The story line is a bit predictable but it's a nice one, sweet and
hilarious in its own right."}\\
(Neg) \textit{``It's a welcome return to form an episode of Snow White, and it turns in a great way."}
\iftoggle{long}{
(Pos) \textit{``The movie's story is based on real life events, and the characters are even more interesting than the original."}\\ 
(Neut) \textit{``Well acted, well acted and flawlessly acted."}\\[0.5em]}{}
\subsection{Examples of Generated Sentences}
All of the examples throughout the paper were generated by the conditioned LM.
Additional examples are available in the supplementary material.

\subsection{Generalization Ability}
\label{sec:eval-generalize}
Finally, we test the ability of the model to generalize: can it generate
sentences for parameter combinations it has not seen in training?
To this end, we removed from the training set the 75,421 sentences which were
labeled as \texttt{theme:plot} and \texttt{personal:true}, and re-trained a
conditioned LM. The trained model did see 336,567 examples of \texttt{theme:plot}
and 477,738 examples of \texttt{personal:true}, but has never seen examples where
both conditions hold together.
We then asked the trained model to generate sentences with these parameter
values. \textbf{100\%} of the generated sentences indeed contained personal pronouns, and
\textbf{82.4\%} of them fit the \texttt{theme:plot} criteria (in comparison, 
a conditioned model trained on \emph{all} the training data managed to fit the \texttt{theme:plot}
criteria in \textbf{97.8\%} of the cases). Some generated sentence examples are:
\\[0.4em]
\noindent
\textit{``Some parts weren't as good as I thought it would be and the acting and script were amazing."}\\[0.4em] 
\textit{``I had a few laughs and the plot was great, but the movie was very predictable."}\\[0.4em]
\noindent
\textit{``I really liked the story and the performances were likable and the chemistry between the two leads is great."}\\[0.4em]
\noindent
\textit{``I've never been a fan of the story, but this movie is a great film that is a solid performance from Brie Larson and Jacob Tremblay.}

\vspace{-0.5em}
\section{Related Work}
\vspace{-0.5em}
\noindent\textbf{In neural-network based models} for language generation, most work focus on content that need to be conveyed in the generated text.
Similar to our modeling approach, \cite{lipton2015capturing, tang2016context} generates reviews conditioned on parameters such as category, and numeric rating scores.
Some work in neural generation for dialog \cite{wen2015semantically,duvsek2016seq2seq,duvsek2016context} condition on a dialog act (``request'', ``inform'') and a set of key,value pairs of information to be conveyed (``price=low, food=italian, near=citycenter''). The conditioning context is encoded either similarly to our approach, or by encoding the desired information as a string and using sequence-to-seqeunce modeling with attention.
\citet{mei2016selective} condition the content on a set of key,value pairs using an encoder-decoder architecture with a coarse-to-fine attention mechanism.
\citet{choiglobally} attempt to 
generate a recipe given a list of ingredients that should be mentioned in the text, tracking the ingredients that were already mentioned to avoid repetitions. \citet{lebretneural} condition on structured information in Wikipedia infoboxes for generating textual biographies. 
\footnote{Recent work by \citet{radford2017learning} trained an unconditioned LSTM language model on movie reviews, and found in a post-hoc analysis a single hidden-layer dimension that allows controling the sentiment of the generated reviews by fixing its value. While intriguuing, it is not a reliable method of deriving controllable generation models.}
These work attempt to control the content of the generated text, but not its style.

In other works, the conditioning context correspond to a specific writer or a group of writers. In generation of conversational dialog, \citet{li2016persona} condition the text on the speaker's identity. While the conditioning is meant for improving the factual consistency of the utterances (i.e., keeping track of age, gender, location), it can be considered as conditioning on stylistic factors (capturing personal style and dialect). 
A recent work that explicitly controls the style of the generated text was introduced by \citet{sennrich2016controlling} in the context of Machine Translation. Their model translates English to German with a feature that encodes whether the generated text (in German) should express politeness.
All these works, with the exception of Sennrich et al condition on parameters that were extracted from meta-data or some database, while Sennrich et al heuristically extracts the politeness information from the training data. Our work is similar to the approach of Sennrich et al but extends it by departing from machine translation, conditioning on numerous stylistic aspects of the generated text, and incorporating both meta-data and heuristically derived properties.

The work of \citet{hu2017controllable} features a VAE based method coupled with a discriminator network that tackles the same problem as ours: conditioning on multiple aspects of the generated text. The Variational component allows for easy sampling of examples from the resulting model, and the discriminator network directs the training process to associate the desired behavior with the conditioning parameters. Compared to our work, the VAE component is indeed a more elegant solution to generating a diverse set of sentences. However, the approach does not seem to be scalable: \citet{hu2017controllable} restrict themselves to sentences of up to length 16, and only two conditioning aspects (sentiment and tense). We demonstrate that our conditioned LSTM-LM appraoch easily scales to naturally-occuring sentence lengths, and allows control of 6 individual aspects of the generated text, without requiring a dedicated discriminator network. The incorporation of a variational component is an interesting avenue for future work. 

\noindent\textbf{In Pre-neural Text Generation} The incorporation of stylistic aspects was discussed from very early on \cite{mcdonald1985computational}. Some works tackling stylistic control of text produced in a rule-based generation system include the works of \citet{power2003generating,reiter2010generating,hovy1987generating,bateman1989phrasing} (see \cite{mairesse2011controlling} for a comprehensive review). Among these, the work of \citet{power2003generating}, like ours, allows the user to control various stylistic aspects of the generated text. This works by introducing soft and hard constraints in a rule-based system.  The work of \citet{mairesse2011controlling} introduce statistics into the stylistic generation process, resulting in a system that allows a user to specify 5 personality traits that influence the generated language.\\
\indent More recent statistical generation works tackling style include \citet{xu2012paraphrasing} who attempt to paraphrase text into a different style. They learn to paraphrase text in Shakespeare's style to modern English using MT techniques, relying on the modern translations of William Shakespeare plays.
\citet{sheikha2011generation} generate texts with different formality levels by using lists of formal and informal words.\\
\noindent\textbf{Finally}, our work relies on heuristically extracting stylistic properties from text. Computational modeling of stylistic properties has been the focus of several lines of study, i.e. \cite{pavlick2016empirical,yang2014detecting,pavlick2015inducing}. Such methods are natural companions for our conditioned generation approach.

\vspace{-0.5em}
\section{Conclusions}
\vspace{-0.5em}
We proposed a framework for NNLG allowing for relatively fine-grained control on different
stylistic aspects of the generated sentence, and demonstrated its effectiveness
with an initial case study in the movie-reviews domain. 
A remaining challenge is providing finer-grained control on the
generated \emph{content} (allowing the user to specify either almost
complete sentences or a set of structured facts) while still allowing the model
to control the style of the generated sentence.

\paragraph{Acknowledgements}
The research was supported by 
the Israeli Science Foundation (grant number 1555/15)
and the German Research Foundation via the
German-Israeli Project Cooperation (DIP, grant DA 1600/1-1).

\bibliography{emnlp2017}
\bibliographystyle{emnlp_natbib}

\clearpage

\section*{Supplementary Materials}

\subsection*{1. Technical Details}

We implement the conditioned LSTM using the
DyNet toolkit.\footnote{https://github.com/clab/dynet}.

\paragraph{Parameters} We use words embedding of size 256 and set the dimension of the model parameters values to 20. Thus, the input vector for the RNN is of size 376 (=256+(20*6)). Our RNN is an LSTM \cite{hochreiter1997long} which was
proven as effective in many NLP tasks. The LSTM's dimension is 1024. The output of the LSTM is fed to an MLP which
is then used for predicting the next word.  The MLP's hidden layer is also of
size 1024.

\paragraph{Training} We use mini-batches of size 16,
and use the Adam optimizer for 10 iterations over the train-set, where 
samples are randomly shuffled before each iteration. We choose the model with
the highest perplexity on the development set.

\paragraph{Generation} When generating texts, we apply \textit{temperature}, which is commonly used to shape the distribution resulting from a
softmax layer, and is defined as follows:  
\begin{equation}
temperature(r_t, \tau) = softmax(\dfrac{r_t}{\tau})
\end{equation}
where $\tau \leq 1$ is the temperature parameter.
The resulting probabilities are nearly the same for high values of $\tau$, while
for low values of $\tau$
the probability of the item with the highest expected reward tends to 1. We set $\tau$ to $0.6$. Then we randomly choose the next token according to the distribution.

\paragraph{Vocabulary}
To cope with a large vocabulary, we follow Sennrich et al.
\shortcite{sennrich2015neural} and use BPE encoding for words. The BPE
(\textit{byte pair encoding}) algorithm \cite{gage1994new} represents an open vocabulary through
a fixed-size vocabulary by encoding rare words as sequences of subword units.
After decoding, subwords units are unified back into single words.
We set the vocabulary size to 30,469.

\section*{2. Categorization to Themes}
The possible values for the \texttt{theme} parameter are: \texttt{plot}, \texttt{acting}, \texttt{production}, \texttt{effects} and \texttt{other}.
We collected common words in advance and assigned them to these categories as follows:

\paragraph{Plot}
story, storytelling, backstory, story-telling, story-line, storyteller, back-story, storytellers, storybook, story-lines, stories, plot, plots, subplot, subplots, plotting, plotline, plotted, sub-plot, sub-plots, plotless, plotlines, plotholes, plot-holes, plot-wise, plot-line, plot-lines, plot-less, script, scripts, scripted, scripting, scripture, scriptwriter, scriptwriting, scriptwriters, unscripted, manuscript, well-scripted, scriptures, manuscripts, written, writing, writer, write, writers, well-written, writes, screenwriter, screenwriters, screenwriting, co-writer, underwritten, co-written, writings, Screenwriter, storyline, storylines, tale, tales, fairytale, fairytales, trilogy, trilogies, trilogía, trilogia, trilogie, screenplay, Screenplay, screenplays, characters, character, characterization, characteristics, characterisation, character-driven, characterizations, characteristic, characterized, scenes, scene, scenery

\paragraph{Acting}
Acting acting, actors, actor, act, acted, actress, acts, actresses, well-acted, actores, acteurs, cast, casting, miscast, casts, casted, well-cast, miscasting, performance, performances, role, roles, played, plays, play, playing, players, player, underplayed, well-played , underplaying

\paragraph{Production}
director, directors, directorial, directores, directora, directoral, directed, directing, production, productions, post-production, pre-production, co-production

\paragraph{Effects}
effects, playlist, songs, song, singer-songwriter, singer/songwriter, songwriting, songwriters, songwriter, music, special-effects, voice, voices, voiced, voiceover, voice-over, voiceovers, voice-overs, visual, visuals, visually, visuales, visualmente, visualize, visualization, visualized, trailer, trailers, sound, sounds, sounded, sounding, soundtrack, soundtracks, cinematography, shot, shots, well-shot\\

\noindent
A sentence is assigned to the theme with the highest number of the matched words in the list. If there are no matched words, the sentence is assigned to the \texttt{other} class.

\section*{3. Examples of Generated Sentences}
These are some examples of sentences that were generated by our model with the corresponding parameters values:

\noindent\rule{8cm}{0.4pt}

\begin{center}
\scalebox{0.8}{
\begin{tabular}{l|l} 
Parameter & Value\\
\hline
Theme & Other    \\
Sentiment &  Negative \\ 
Professional &  False   \\
Personal & True \\
Length & 11-20 words    \\
Descriptive & True   \\
\end{tabular}}
\end{center}

\begin{itemize}
\item ``My biggest problem with the whole movie though is that there is nothing new or original or great in this film."
\item ``There are some funny parts but overall I didn't like the first few funny parts, but overall pretty decent ."
\item ``My biggest problem with the movie was the fact that is managed to use the same exact same well-written line."
\item ``Ultimately, I can honestly say that this movie is full of stupid stupid and stupid stupid stupid stupid stupid."
\item Good but a little bit slow and boring, I was looking forward to seeing this movie with my parents.
\end{itemize}

\noindent\rule{8cm}{0.4pt}

\begin{center}
\scalebox{0.8}{
\begin{tabular}{l|l} 
Parameter & Value\\
\hline
Theme & Other    \\
Sentiment &  Negative \\ 
Professional &  False   \\
Personal & False \\
Length & 11-20 words    \\
Descriptive & True   \\
\end{tabular}}
\end{center}

\begin{itemize}
\item ``A little bit of a predictable and boring romantic comedy with a few funny moments but overall pretty entertaining."
\item  ``With such a great premise ,Escape From Tomorrow is pretty damn terrible, horrible, and no exception."
\item ``The first half is lazy and stupid, but there's a handful of funny moments which are pretty decent."
\item ``There's no denying the fact that this movie is such a horrible movie with a few bad moments."
\item
``The last part of the movie just let me down, but the whole thing is pretty good."
\item
``It is a little difficult to follow, but this is a rare right choice for the respective aspects of film."
\item
``My biggest issue is that the first half is pretty boring, plodding, and too obvious to be honest."
\end{itemize}

\noindent\rule{8cm}{0.4pt}

\begin{center}
\scalebox{0.8}{
\begin{tabular}{l|l} 
Parameter & Value\\
\hline
Theme & Plot    \\
Sentiment &  Positive \\ 
Professional &  False   \\
Personal & False \\
Length & 11-20 words    \\
Descriptive & False   \\
\end{tabular}}
\end{center}

\begin{itemize}
\item ``The movie's story is based on real life events, and the characters are even more interesting than the original."
\item ``The story is great, the most fun and the most great film you'll see in a long time."
\item ``The story is the best part of the movie, it's a lot of fun to watch."
\item ``It 's a touching story that will keep you on the edge of your seat the whole time ! ! !"
\item ``The story was not quite as good as the first one but it had a pretty good twist ending."
\item ``It's a story that doesn't take itself too seriously, but it's a surprisingly good film."
\item ``The movie is a perfect love story that leaves you with a smile on your face for the great effect."
\item 
``The characters were great and the storyline had me laughing out loud at the beginning of the movie."
\item ``The two main characters are a bit of a stretch, but this movie is still very well done."
\item ``The film is definitely a unique thing to follow, and some of the characters are just a bit too complicated."
\end{itemize}

\noindent\rule{8cm}{0.4pt}

\begin{center}
\scalebox{0.8}{
\begin{tabular}{l|l} 
Parameter & Value\\
\hline
Theme & Other    \\
Sentiment &  Positive \\ 
Professional &  True   \\
Personal & False \\
Length & 11-20 words    \\
Descriptive & False   \\
\end{tabular}}
\end{center}

\begin{itemize}
\item
``This is a must see for fans of Bergman's `` American Dream '', `` The Roots ''."
\item
``The film's ultimate pleasure if you want to fall in love with the ending, you won't be disappointed"
\item
``One of the most inspirational films I've seen in years, hence the most influential, the best."
\item
``The monetary system is a bit too intelligent and at times implanted in the middle of the film."
\item
``The film doesn't seem to be anything more than a dozen other films with a sophisticated vigor."
\item
``The film's late 19th century Denmark's history is a feast for the eyes and the laughter on the screen."
\item
``
The film's simple, and a refreshing take on the complex family drama of the regions of human intelligence."
\end{itemize}

\noindent\rule{8cm}{0.4pt}

\begin{center}
\scalebox{0.8}{
\begin{tabular}{l|l} 
Parameter & Value\\
\hline
Theme & Acting    \\
Sentiment &  Neutral \\ 
Professional &  False   \\
Personal & False \\
Length & 21-40 words    \\
Descriptive & False   \\
\end{tabular}}
\end{center}

\begin{itemize}
\item
``It's a shame to see the actors playing the corporation's name or the West Memphis Three in the theater, but he still manages to do with a one of a better black comedy."
\item
``In the end, the film is a brilliantly acted and terrifically paced adaptation of the dystopian book, but the direction is completely questionable as its subject matter gives us a sense of realism which is a brilliant thing" 
\item
``A good performance from James Franco and Josh Brolin, Peter Dinklage ( played by Michael Douglas ), his wife and the city of the mayor, who develops into a newly dramatic movie."
\end{itemize}

\end{document}